\documentclass{article}

\usepackage{arxiv}
\usepackage{amsmath}
\usepackage[section]{algorithm}
\usepackage{pifont}
\usepackage{algpseudocode}
\usepackage{algorithmicx}%
\usepackage{listings}
\usepackage{booktabs}
\usepackage{tabularx}
\usepackage{multirow}
\usepackage{hyperref}%
\usepackage[utf8]{inputenc} % allow utf-8 input
\usepackage[T1]{fontenc}    % use 8-bit T1 fonts
\usepackage{hyperref}       % hyperlinks
\usepackage{url}            % simple URL typesetting
\usepackage{booktabs}       % professional-quality tables
\usepackage{amsfonts}       % blackboard math symbols
\usepackage{nicefrac}       % compact symbols for 1/2, etc.
\usepackage{microtype}      % microtypography
\usepackage{lipsum}
\usepackage{graphicx}
\graphicspath{ {./images/} }

\newcommand{\cmark}{\ding{51}}%
\newcommand{\xmark}{\ding{55}}%

\title{BoostTrack++: using tracklet information to detect more objects in multiple object tracking}

\author{
 Vukašin Stanojević \\
  Faculty of Sciences and Mathematics\\
  University of Niš\\
  \texttt{vukasin.stanojevic@pmf.edu.rs} \\
   \And
 Branimir Todorović \\
 Faculty of Sciences and Mathematics\\
  University of Niš\\
  \texttt{branimir.todorovic@pmf.edu.rs} \\
}
\date{}
\begin{document}
\maketitle
\begin{abstract}
Multiple object tracking (MOT) depends heavily on selection of true positive detected bounding boxes.
However, this aspect of the problem is mostly overlooked or mitigated by employing two-stage association and
utilizing low confidence detections in the second stage.
Recently proposed BoostTrack attempts to avoid the drawbacks of multiple stage association approach and use
low-confidence detections by applying detection confidence boosting.
In this paper, we identify the limitations of the confidence boost used in BoostTrack and propose a
method to improve its performance.
To construct a richer similarity measure and enable a better selection of true positive detections,
we propose to use a combination of shape, Mahalanobis distance and novel soft BIoU similarity.
We propose a soft detection confidence boost technique which calculates new confidence scores based on the similarity measure and the previous confidence scores,
and we introduce varying similarity threshold to account for lower similarity measure between detections and tracklets which are not regularly updated.
The proposed additions are mutually independent and can be used in any MOT algorithm.

Combined with the BoostTrack+ baseline, our method achieves near state of the art results on the MOT17 dataset and new state of the art HOTA and IDF1 scores on the MOT20 dataset.

The source code is available at: \url{https://github.com/vukasin-stanojevic/BoostTrack}.
\end{abstract}

% keywords can be removed

    \section{Introduction}
    Multiple object tracking (MOT) is an important and active topic in computer vision.
    The main applications include human-robot interaction ~\cite{wengefeld2019multi}, autonomous driving ~\cite{wang2021tracking} and surveillance ~\cite{jha2021real},
    but it can also be applied to analyse sports videos ~\cite{cui2023sportsmot}, track animals ~\cite{zhang2023animaltrack} and even in medicine ~\cite{li2024research}.
    Given a video with multiple objects of interest (e.g. pedestrians), the aim is to construct a
    trajectory for each object.
    More specifically, for each frame, every object of interest should be detected
    and assigned an ID, which should not change during the video even if the object
    is not present in every frame (e.g. it can be occluded).
    MOT can be solved offline by processing the entire video, or online by processing one frame at a time without considering the future frames.
    Online MOT solutions have wider applications and can be used for real-time tracking in autonomous driving or surveillance systems.
    Among the online methods, tracking by detection (TBD) methods show the best performance.
    In TBD paradigm, MOT is solved in two steps: detection, which outputs a set of detected bounding boxes; and association
    in which detected bounding boxes should be associated (matched) with currently tracked objects.
    Hungarian algorithm ~\cite{kuhn_hungarian_1955} is a typical choice for matching between newly detected bounding boxes and existing tracklets, i.e. current states of tracked trajectories.
    The cost matrix used by the algorithm can be constructed by combining different similarity measures such as intersection over union
    (IoU), Mahalanobious distance ~\cite{mahalanobis_generalized_2018} or appearance similarity (cosine similarity between visual embedding vectors).

    Before constructing a cost matrix, some method of filtering out false positive detections should be applied.
    This is usually achieved by discarding detections with confidence scores below a specified threshold.
    This simple logic results in discarding some true positive detections also.
    To mitigate this, ByteTrack ~\cite{zhang_bytetrack_2022} employed a two-stage association in which low-confidence detections and unmatched tracklets
    are used in the second association stage.
    Two-stage association became the standard in TBD MOT (e.g. standard benchmark
    methods that adopted two-stage association include ~\cite{aharon_bot-sort_2022, maggiolino_deep_2023, du_strongsort_2023}).
    % and the multiple stage association is further expanded in some recent works (e.g. ~\cite{li_motion_2024, yang2024local} used three-stage and ~\cite{meng2023localization} used four-stage association).
    However, multiple stage association methods can introduce identity switches (IDSWs) ~\cite{stadler_improved_2023}.
%     IDSW can occur in case when two tracklets are hardly distinguishable but their respective detection
%    boxes are to be associated in different stages, a mismatch in one stage forces the mismatch in the other, which could
%    have been avoided if both detected bounding boxes were associated in the same stage.

    Recently, BoostTrack ~\cite{stanojevic_boost_2024} used a one-stage association combined with strategies to increase (boost) the detection confidence of some
    detected bounding boxes before discarding detections with low confidence scores.
    % The idea is to use more true positive detections that would otherwise be discarded and still perform simple one-stage association.
    In confidence boost based on detection of likely objects (DLO), it used intersection over union (IoU) between
    all detected bounding boxes and all tracklets to discover the low-confidence detections with high overlap with existing tracklet.
    BoostTrack boosted the confidence score of these detections assuming they correspond to currently tracked objets,
    which improved the tracking performance in crowded scenes with frequent occlusions (namely, on the MOT20 dataset).
    However, the method resulted in an increased number of new IDs, and identity switches, which indicates a more sophisticated method is required.
    In this paper, we extend the idea of DLO confidence boost.

    In \cite{stanojevic_boost_2024}, shape similarity and Mahalanobis distance are used to improve ("boost") the similarity measure to
    reduce ambiguity arising from using IoU only.
    This improved association performance.
    However, if a particular similarity measure improves association, it should also improve the selection of low-confidence true positive detections.
    We extend the idea of buffered IoU (BIoU) from ~\cite{yang_hard_2023} and create a novel soft BIoU similarity measure,
    which we use jointly with Mahalanobis distance and shape similarity from ~\cite{stanojevic_boost_2024} to discover true positive low-confidence detections.

% First, instead of using only the IoU to find such detections, we use a combination
% of other similarity measures, namely the Mahalanobis distance and shape similarity used in \cite{boostTrack}.
    DLO confidence boost used in \cite{stanojevic_boost_2024} uses only the similarity (IoU) between a detection and a tracklet
    to compute the boosted similarity.
    If similarity is above the certain threshold, the detection will be used.
    This means that the detection whose confidence is close to zero
    and the detection that has a confidence score close to the threshold (slightly below it) will be equally treated in the
    DLO boost procedure, i.e., both require the same high similarity to get the boost.
    We introduce a soft detection confidence boost which takes into account the original detection confidence score
    and solves the described problem.

% Second, instead of boosting the confidence score based solely on whether the similarity measure surpasses some threshold
% value, which BoostTrack implicitly do, we use soft detection confidence boost which takes into account original detection
% confidence score.
    Another weakness of the BoostTrack DLO confidence boost technique is that
    it treats all tracklets equally when deciding if IoU between a tracklet and a detection is high enough.
    However, the tracklets that are not recently updated (i.e. not matched with a detected bounding box for multiple subsequent frames due to occlusion, detector or association failure) usually have relatively low IoU with the corresponding
    detections once they are matched.
    Setting a threshold for "high enough" similarity should be tracklet specific and depend on the number of
    frames since the last time the tracklet was updated.
    For example, if a tracklet was not matched for 30 frames, IoU of 0.8 can be considered very high.
    We attempt to solve this problem by introducing varying threshold based on the number of frames since the last update of a given tracklet.

    Each of the proposed modifications is independent to the others and can be combined and used in any TBD MOT algorithm.
    We perform a detailed ablation study on MOT17 ~\cite{milan_mot16_2016} and MOT20 ~\cite{MOTChallenge20} validation sets
    to show the effectiveness of each component.
    We successfully reduced the number of new IDs and identity switches, while not only retaining baseline tracking performance
    but surpassing it.

\begin{figure}[h!t]
        \centering
        \includegraphics[width=.99\linewidth]{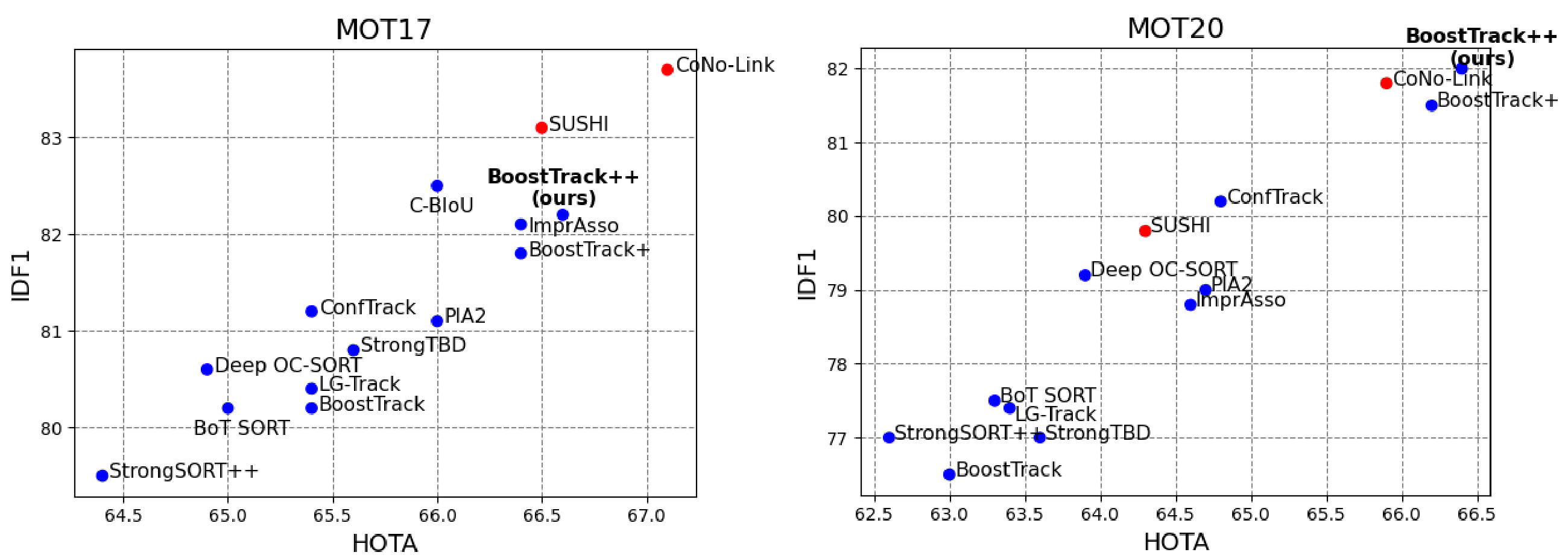}
        \caption{Results of HOTA and IDF1 metrics on MOT17 (left) and MOT20 (right) test sets. We display offline methods as red circles.}
        \label{fig:grafik}
    \end{figure}

    We name the MOT system that combines BoostTrack+ baseline with the proposed additions BoostTrack++.
    Among online trackers, BoostTrack++ ranks first in HOTA score on the MOT17 test set.
    On MOT20, BoostTrack++ ranks first in HOTA and IDF1 scores among all trackers (see figure ~\ref{fig:grafik}).

    In summary, we make the following contributions:
    \begin{itemize}
        \item We introduce a novel soft buffered IoU (soft BIoU) similarity measure.
        \item We use the combination of Mahalanobis distance, shape and soft BIoU similarity to calculate the more sophisticated
        similarity measure for detecting the likely objects.
        \item We use soft detection confidence boost which uses the original detection confidence score for calculating the boosted confidence.
        \item We introduce varying boost threshold based on the number of time steps since the last update of a given tracklet.
        \item Added to the BoostTrack+ baseline, our BoostTrack++ method achieves near state of the art results on MOT17 and sets new state of the art on MOT20 dataset.
    \end{itemize}

    The rest of the paper is structured as follows:
    In section ~\ref{sc:preliminaries}, we give a brief introduction to the TBD MOT approach and cover the
    multiple-stage association methods, which is the standard for dealing with low-confidence detections.
    Since our method extends the BoostTrack, we give a short overview of BoostTrack in subsection ~\ref{ssc:boosttrack}.
    We discuss the drawbacks of the DLO confidence boost from BoostTrack in section ~\ref{sc:limitations}.
    Section ~\ref{sc:methods} covers the proposed method.
    First, in subsection ~\ref{ssc:sbiou} we introduce our novel soft buffered IoU similarity measure.
    In the next subsection, we introduce the average similarity measure for the detection confidence boost,
    after which, in subsections ~\ref{ssc:softDLO} and ~\ref{ssc:varyingTh} we introduce our soft detection confidence
    boost and varying threshold techniques, respectively.
    In subsection ~\ref{ssc:methodAll} we introduce an algorithm which combines all of the proposed additions.
    In section ~\ref{sec:experiments}, we provide the results of the ablation study and compare the proposed method
    with other methods on MOT17 and MOT20 test sets.
    We conclude the paper with section ~\ref{sec:conclusion}.

    \section{Preliminaries}
    \label{sc:preliminaries}
    \subsection{Tracking by detection}
    One of the dominant approaches in MOT is to follow the TBD paradigm.
    In TBD, the MOT problem is solved frame-by-frame by performing two steps of processing for every frame: detection and association.
    Given a frame, the detector model (e.g. YOLOX ~\cite{ge2021yolox}) outputs the set of detected bounding boxes
    $D=\{D_1,D_2, \ldots , D_n\}$ with the corresponding confidence scores $c_d =\{c_{d_1}, c_{d_2}, \ldots , c_{d_n}\}$.
    Tracking module (e.g. Kalman filter ~\cite{kalman_new_1960}) is needed to predict the state of currently tracked objects, i.e. tracklets
    $T = \{T_1, T_2, \ldots T_m\}$, based on the past states (one state in the case of Kalman filter).

    If Kalman filter is used, the state of the object usually consists of object location and size, and the corresponding
    velocities.
    In this paper, we use the same settings for the Kalman filter as in ~\cite{stanojevic_boost_2024}.

    Linear state space model of tracking is given as:
    \begin{equation}
        \begin{aligned}
            x_k &= F \cdot x_{k-1} + q_k \quad \text{(dynamic equation)},\\
            y_k &= H\cdot x_k + r_k \quad \text{(observation equation)},
        \end{aligned}
    \end{equation}
    where $q_k$ and $r_k$ are process and observation noises, respectively (we use the same noise setting as in ~\cite{stanojevic_boost_2024}).
    The state $x_k$, state transition matrix $F$, and observation matrix $H$ are given by:
    \begin{equation}
        \label{eq:kalmanSetup}
        \mathbf{x_k} = [u, v, h, r, \dot{u}, \dot{v}, \dot{h}, \dot{r}]^{\mathsf{T}},\quad
        F =
%            \begin{bmatrix}
%                  1 & 0 & 0 & 0 & 1 & 0 & 0 & 0 \\
%                  0 & 1 & 0 & 0 & 0 & 1 & 0 & 0 \\
%                  0 & 0 & 1 & 0 & 0 & 0 & 1 & 0 \\
%                  0 & 0 & 0 & 1 & 0 & 0 & 0 & 1 \\
%                  0 & 0 & 0 & 0 & 1 & 0 & 0 & 0 \\
%                  0 & 0 & 0 & 0 & 0 & 1 & 0 & 0 \\
%                  0 & 0 & 0 & 0 & 0 & 0 & 1 & 0 \\
%                  0 & 0 & 0 & 0 & 0 & 0 & 0 & 1
%        \end{bmatrix}
        \begin{bmatrix}
            I_4 & I_4\\
            \mathbf{0_{4\times4}} & I_4
        \end{bmatrix},\quad
        H =
        \begin{bmatrix}
            I_4 & \mathbf{0_{4\times4}}
        \end{bmatrix},
    \end{equation}
    where $u$, $v$, $h$ and $r$ represent the coordinates of the object center, the height of the object and the ratio of the object's width and height, respectively.
    By $\dot{u}, \dot{v}, \dot{h}, \dot{r}$ we denote their corresponding velocities.
    When a tracklet is associated with detected bounding box $D_i$, that bounding box is used as the observation in Kalman update step.

    Tracklets and detections can be matched using the Hungarian algorithm ~\cite{kuhn_hungarian_1955}.
    The cost matrix $C$ for the matching can be constructed
    simply as $C = -1 \cdot S(D, T)$, where $S(D, T)$ is the similarity matrix between the detections $D$ and tracklets $T$.
    In the simplest case, $S=\text{IoU}$.

    Looking at TBD logic with more details, before constructing the cost matrix, a selection of "reliable"
    detected bounding boxes is required, i.e. false positive detections should be filtered out.
    The simplest method is thresholding - deciding which detections
    will be used based on whether the confidence score is greater than a specified threshold value $\tau$.
    The matching between a detected bounding box $D_i$ and a tracklet $T_j$ is admissible if $S(D_i, T_j)>\tau_S$, where $\tau_S$
    represents a minimal required similarity (e.g. $\tau_S=0.3$).
    Unmatched detections are used to initialize new tracklets\footnote{In some implementations, there is a separate threshold for
    tracklet creation $\tau_{init}, \tau_{init}>\tau$, and only the detections with confidence scores greater than $\tau_{init}$
        can be used for tracklet initialization.}.
%    Sometimes the detector can output false positive detection with high confidence score.
%    However, this can rarely happen for many consecutive frames.
%    To mitigate this issue, an object state is reported only when the object was successfully
%    matched for at least $n_{match}$ consecutive frames.

    \subsection{Multiple-stage association methods}
    Not all low-confidence detections are false positives.
    The dominant approach to using low-confidence detections (and utilising the true positive detections that would have been discarded otherwise)
    is using a two-stage association which was first introduced in ByteTrack  ~\cite{zhang_bytetrack_2022}.
    ByteTrack uses thresholding but does not discard low-confidence detections.
    Instead, the unmatched tracklets are matched with low-confidence detections in the second association stage.
    Followed by ByteTrack, the two-stage association approach became the standard in TBD MOT (e.g. it is used in
    ~\cite{ren2024rlm,feng2023towards,li2024lightweight,wang2024smiletrack,aharon_bot-sort_2022,bui2024camtrack,li_simpletrack_2022}).
    Some tracking methods expanded this logic further.
    Tracking algorithms in  ~\cite{li_motion_2024, yang2024local} performed three-stage association,
    LG-Track used four-stage association ~\cite{meng2023localization}, while some other
    methods used multiple-stage association based on tracklet last update ~\cite{wojke_simple_2017}, or similarity between
    detections and tracklets ~\cite{nasseri_fast_2022}.

    However, it is demonstrated in ~\cite{stadler_improved_2023} that two-stage association can introduce identity switches.
    The same applies to any multiple-stage association strategy which does not consider all tracklet-detection pairs in the
    same stage.

    \subsection{BoostTrack}
    \label{ssc:boosttrack}
    BoostTrack is a TBD system built upon SORT ~\cite{bewley_simple_2016}, and it tackles three problems in MOT:
    finding a simple and better similarity measure to improve association performance,
    the selection of true positive detections, and using a simple one-stage association to avoid identity switches to which
    all multiple-stage association techniques are prone to ~\cite{stadler_improved_2023}.
    To improve the association performance, BoostTrack uses three additions to IoU (to "boost" the original IoU similarity) and
    defines the overall similarity measure between a detected bounding box $D_i$ and a tracklet $T_j$, $S(D_i, T_j)$, as

    \begin{equation}
        \label{eq:symBoostEq}
        S(D_i, T_j) = \text{IoU}(D_i, T_j) + \lambda_{IoU} \cdot c_{i,j} \cdot \text{IoU}(D_i, T_j)
        + \lambda_{MhD} \cdot S^{MhD}(D_i, T_j) + \lambda_{shape} \cdot S^{shape}(D_i, T_j),
    \end{equation}
    where $c_{i,j}$, $S^{MhD}$ and $S^{shape}$ represent detection-tracklet confidence of detection $D_i$ and tracklet $T_j$,
    Mahalanobis distance similarity and shape similarity, respectively.

    To select more true positive detections for the first (and only) association stage,
    BoostTrack uses two detection confidence boosting techniques.
    Before discarding low-confidence detections, it increases the confidence score of bounding boxes which \textit{should} be true positives.
    One technique relies on Mahalanobis distance to find outliers.
    Assumption is that the detections which do not correspond to any of the currently
    tracked objects are not false positives, but rather detections corresponding to previously undetected objects.
    The other technique, detecting the likely objects, uses IoU between a detection and all the tracklets.
    If low-confidence detection has a high IoU with some tracklet, the detected bounding box is considered to correspond to that tracklet,
    and its detection confidence score should be increased.
    Equation ~\ref{eq:iouConfBoost} shows the expression for increasing detection confidence score $c_{d_i}$ for some detected
    bounding box $D_i$.
    \begin{equation}
        \label{eq:iouConfBoost}
        \hat{c}_{d_i} = \text{max}\big(c_{d_i}, \beta_c\cdot\max_{j}(\text{IoU}(D_i, T_j))\big).
    \end{equation}
    Hyperparameters $\beta_c$ and $\tau$ implicitly define the IoU threshold required for a low-confidence detection $D_i$ to
    surpass $\tau$.
    Values used in ~\cite{stanojevic_boost_2024} correspond to $0.923$ and $0.8$ for datasets MOT17 ~\cite{milan_mot16_2016} and
    MOT20 ~\cite{MOTChallenge20}, respectively.

    The authors also proposed BoostTrack+ method which uses appearance similarity (between embedding vectors) in addition to
    similarity measures used in ~\ref{eq:symBoostEq}.
    BoostTrack+ outperforms BoostTrack at the expanse of longer computation time.

    In table ~\ref{tb:methods}, we summarize the methods used as baselines in this paper.
    \begin{table}[h!t]
        \scriptsize
        \center
        \caption{Summary of various baseline methods.}
        \begin{tabular}{c|c|c|c|c}
            \toprule
            Method & Description & Pros & Cons & Improvements over prev. method \\ % \cmidrule{1-11}
            \midrule
            \multirow{3}{*}{SORT} & Tracker that uses IoU & One-stage association, & Poor selection of &  \multirow{3}{*}{/} \\
            &as similarity measure & real-time performance &  true-positive detections, \\
            & & & poor tracking performance &\\
            \midrule
            \multirow{4}{*}{BoostTrack} & Tracker focused on & One-stage association, & Lacks strong cues & \multirow{4}{*}{Improved overall performance} \\
            & handling unreliable detections & uses all detections, & such as visual & \\
            & and improving association & real-time performance & features embedding, & \\
            & & & introduces IDs and IDSWs & \\
            \midrule
            \multirow{3}{*}{BoostTrack+} & Tracker focused on & One-stage association, & Cannot operate in & \multirow{3}{*}{Improved tracking performance} \\
            & handling unreliable detections & uses all detections, & real time in crowded scenes,& \\
            & and improving association & & introduces IDs and IDSWs & \\
            \bottomrule
        \end{tabular}
        \label{tb:methods}
    \end{table}

    \subsection{Buffered IoU}
    \label{ssc:biou}
    To account for irregular motions, Buffered IoU (BIoU) is introduced in ~\cite{yang_hard_2023}.
    If the predicted state is inaccurate (due to irregular motion), the IoU between the predicted bounding box and the detected
    bounding box will be low (possibly 0) which makes association difficult or even impossible.
    The authors proposed to scale (add "buffers" to) the detected and the predicted bounding boxes, i.e. tracklets.
    Let $o=(x, y, w, h)$ be the original detection (or tracklet), where $(x, y)$ represents the top-left coordinate of the bounding box,
    and $w$ and $h$ its width and height, respectively.
    The authors propose to use scaled detection $o_b = (x-bw, y-bh, w+2bw, h+2bh)$, where $b\geq0$ is the scale parameter.
    More specifically, they performed two-stage association: in the first stage they used small scale parameter $b_1$,
    and in the second stage they associated remaining tracklets and detections using larger scale parameter $b_2$.
    The logic behind two-stage association is that the unmatched tracklets are more difficult to match and require larger
    bounding boxes for successful matching.

    \section{Limitations of confidence boost based on detection of likely objects}
    \label{sc:limitations}
    One of the contributions of BoostTrack ~\cite{stanojevic_boost_2024} is
    the DLO confidence boost technique.
    The effectiveness of this technique is most notable in the case of the MOT20 dataset,
    where the DLO boost increased MOTA\footnote{We discuss the metrics used in subsection ~\ref{ssc:metrics}.} score
    by 4.8\% (see table ~\ref{tb:dlomot20}).

    To achieve the best results, a different value of hyperparameter $\beta_c$ (see equation ~\ref{eq:iouConfBoost}) had to be
    specified depending on the dataset ($0.65$ for MOT17 and $0.5$ for MOT20), which is one of the limitations of the DLO confidence boost.

    The most important issue is the introduction of new IDs, which should not happen if the detections with boosted
    confidence truly correspond to existing tracklets.
    Ideally, a DLO confidence boost should produce no new IDs or result in identity switches (IDSWs).
    This issue is best illustrated by results on the MOT20 validation set which we present in table ~\ref{tb:dlomot20}.
    On the MOT17 validation set (which has only 339 ground-truth IDs, as opposed to the MOT20 validation set which contains 1418 IDs) the DLO boost is less significant.

    \begin{table}[h!t]
        \center
        \caption{Influence of DLO confidence boost on MOT20 validation set for various baseline methods.}
        \begin{tabular}{c|ccc|cc}
            \toprule
            Method & HOTA & MOTA & IDF1 & IDSWs & IDs \\ % \cmidrule{1-11}
            \midrule
            SORT & 56.65 & 69.91 & 73.6 & 1127 & 1894 \\
            SORT + DLO & 58.58 & 73.28 & 75.09 & 1259 (+132) & 2045 (+151) \\
            \midrule
            BoostTrack - DLO & 61.39 & 77.02 & 77.23 & 803 & 1867 \\
            BoostTrack & 61.74 & 77.46 & 77.45 & 898 (+95) & 2008 (+141) \\
            \midrule
            BoostTrack+ - DLO & 62.59 & 77.16 & 79.29 & 730 & 1852 \\
            BoostTrack+ & 62.58 & 77.7 & 78.93 & 794 (+64) & 2019 (+167) \\
            \bottomrule
        \end{tabular}
        \label{tb:dlomot20}
    \end{table}

    If we take SORT ~\cite{bewley_simple_2016} as the most basic and representative baseline, we notice that the increase in
    main MOT metrics (the left side of table ~\ref{tb:dlomot20}), comes at the price of increased IDs (+8\%) and IDSWs (+12\%).

    As the idea behind the DLO confidence boost is to use the detections with similarity (IoU) above a certain threshold,
    two possible reasons for the method failure and places for improvement are the similarity measure used and the specified threshold.

    Several papers have discussed the limitations of IoU used as a similarity measure for association and used various IoU modifications or other motion cues
    to construct a richer similarity measure (e.g. ~\cite{li_simpletrack_2022,cao_observation-centric_2023,yang2024hybrid,nasseri_fast_2022,stanojevic_boost_2024,morsali2024sfsort}).
    If using IoU alone can cause ambiguities and IDSWs in association, it is also unreliable to base the DLO confidence boost solely on IoU.
    Using a richer similarity measure improves association and enables the discard of falsely associated detection-tracklet pairs.
    Since DLO confidence boost relies on calculating similarity, a richer similarity measure should also improve the selection of
    true positive detections.

    On the other hand, using a fixed threshold (for a given dataset) for the DLO confidence boost has two drawbacks.

    First, it does not take into account the original confidence score.
    A detection with a higher confidence score is more likely to be a true positive.
    If a detection has a confidence score slightly below the threshold $\tau$,
    it should require only a mild conformation in similarity with existing the tracklet to boost its confidence above the $\tau$.

    Second, looking from a tracklet perspective, what can be considered a "high" similarity (e.g. IoU) between a detected bounding box?

    As in BoostTrack ~\cite{stanojevic_boost_2024}, we use the Kalman filter as the tracking module.
    The quality of Kalman filter predictions reduces when the tracklet does not get matched with a detection,
    i.e. when we miss the observation and do not execute the Kalman update step.
    Error covariance prediction in step $t$, $\hat{P}_t$, is calculated as
    $\hat{P}_t = F \cdot P_{t-1} F^\mathsf{T} + Q$ (we provide the definition of the state $x$ and state transition matrix $F$ in equation ~\ref{eq:kalmanSetup}).

    The estimate of the variance of $x \in \{u, v, h, r\}$, after not executing Kalman update for $n$ steps (frames),
    if we omit $Q$ for simplicity, becomes:
    \begin{equation}
        \label{eq:covError}
        \hat{\text{var}}(x)_{t+n} = \text{var}(x)_{t} + n^2\cdot \text{var}(\dot{x})_{t}.
    \end{equation}
    The variance of the predictions increases quadratically with the increase of $n$\footnote{Equality ~\ref{eq:covError} follows directly from substituting $F$ and $x$
        from equation ~\ref{eq:kalmanSetup} and applying prediction for $n$ steps.}.

    \begin{figure}[h!t]
        \centering
        \includegraphics[width=.9\linewidth]{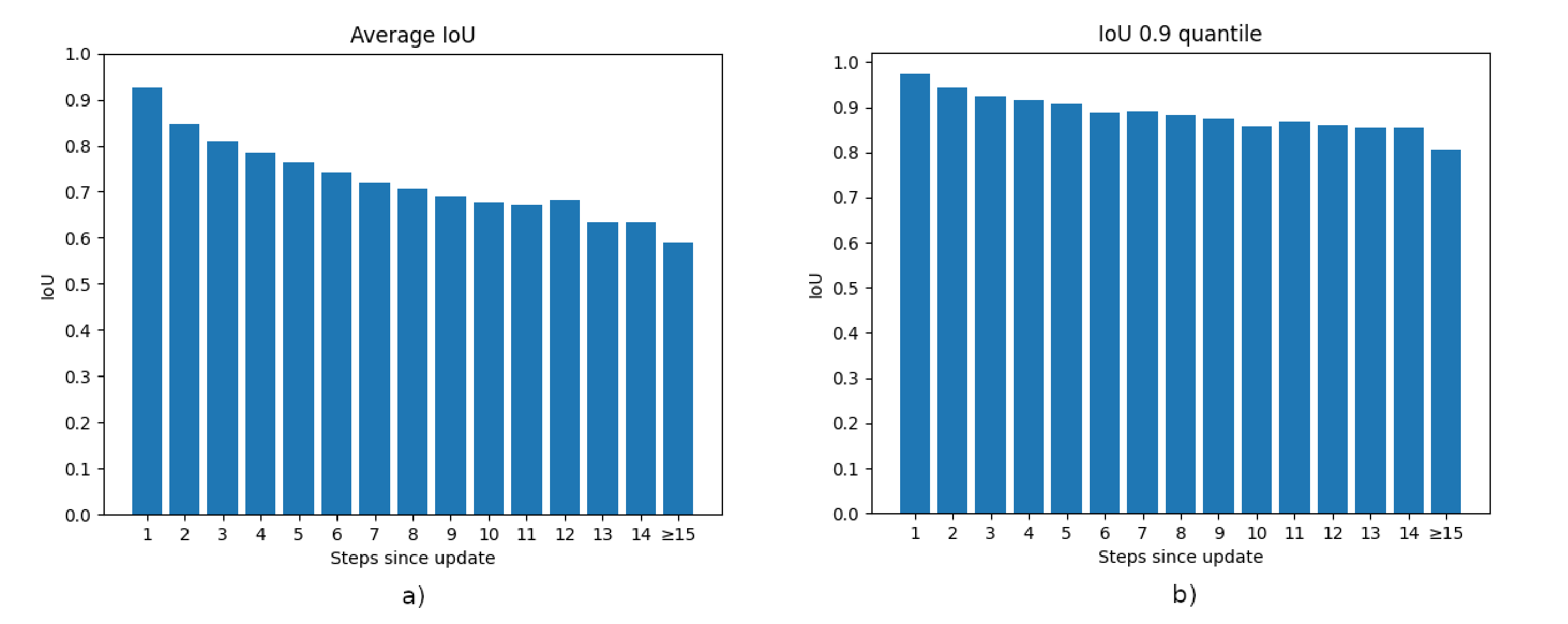}
        \caption{IoU statistics calculated on MOT17 and MOT20 validation sets for tracklets with different numbers of steps since the last update.
        a) Average IoU b) 0.9 quantile of IoU values. }
        \label{fig:iou1}
    \end{figure}

    We empirically verified that the IoU value between a detection and a tracklet decreases as the number of steps since the last
    successful match (last position update) of the given tracklet increases.
    The left side of the figure ~\ref{fig:iou1} shows the average IoU value (calculated on MOT17 ~\cite{milan_mot16_2016} and
    MOT20 ~\cite{MOTChallenge20} validation sets\footnote{For more information on datasets and implementation details, see subsections \ref{ssc:metrics} and \ref{ss:implDetails}.}) depending on the number of frames since the last update,
    while the right side of the same figure shows values of $0.9$ quantiles.
    The $0.9$ quantile gives us information about what can be considered "high" IoU
    depending on the number of steps since the last update of the corresponding tracklet.
    As the number of steps since the last update increases, we can observe a decrease in average IoU,
    and, more importantly, a decrease in $0.9$ quantile.

    The DLO confidence boost method should account for this decrease and use different tracklet-specific thresholds based on the number
    of frames since the last successful match of a given tracklet.

    The data from table ~\ref{tb:dlomot20} and the discussion presented indicate that a more sophisticated DLO
    confidence boost is needed, which we attempt to offer in the following section.

    \section{Proposed methods}
    \label{sc:methods}

    \subsection{Soft BIoU}
    \label{ssc:sbiou}
    BIoU, introduced in ~\cite{yang_hard_2023}, enables to match detection-tracklet pairs which are (due to the inaccurate prediction)
    too separated and have low or 0 IoU, by expanding bounding boxes.
    As such, BIoU could also be used for DLO confidence boost.
    However, BIoU is designed to be used in a two-stage association and cannot be used directly.
    In ~\cite{yang_hard_2023}, the tracklets are, through two-stage association, split into two groups and the corresponding bounding
    boxes are scaled using hyperparameters $b_1$ and $b_2$ depending on the group.

    A trivial way of using BIoU in one-stage association setup would be to use single scale value $b$.
    However, not only that "easy" and "difficult" tracklets require different $b$ values,
    but not all "easy" tracklets are equally easy to match, and not all "difficult" tracklets are equally difficult.

    Every predicted bounding box should be enlarged to the extent which is proportional to the uncertainty in the quality of the prediction.
    In ~\cite{stanojevic_boost_2024}, tracklet confidence is defined and used as a measure of reliability of the predicted tracklet state.
    We propose to use tracklet confidence to calculate tracklet specific scale without the need for a two-stage association.
    However, if we use different scales for different tracklets and comparing with detections in one stage,
    we cannot simply scale the tracklets and detections by the same parameter $b$ and calculate IoU between all pairs, as
    done in ~\cite{yang_hard_2023}.
    Every detection-tracklet pair $(D_i, T_j)$ should use a specific scale based on tracklet confidence, $c_{t_j}$, value.

    Let $\mathbf{o} \rightarrow s$ be detection $\textbf{o}$ scaled by $s$, defined (as in ~\cite{yang_hard_2023}) as
    \begin{equation}
        \label{eq:scaledO}
        \mathbf{o} \rightarrow s = (x-sw, y-sh, w+2sw, h+2sh),
    \end{equation}
    where $(x, y), w$, and $h$ represent top-left corner of the bounding box, its width and height, respectively.
    We define soft BioU (SBioU) between $D_i$ and $T_j$ as
    \begin{equation}
        \label{sbiou}
        \text{SBIoU}(D_i, T_j) = \text{IoU}\big(D_i \rightarrow \frac{1-c_{t_j}}{4}, T_j \rightarrow \frac{1-c_{t_j}}{2}\big),
    \end{equation}
    for $i \in \{1, 2, \ldots , n\}$, $j \in \{1, 2, \ldots , m\}$.
    As the tracklet confidence decreases, the scale increases.
    To calculate the scaling parameter, we divide $1-c_{t_j}$ by $2$ to match the range of scale values from ~\cite{yang_hard_2023}
    (for $c_{t_j}=0.0$ it increases scale to $b_2=0.5$).
    When scaling the detection box, we use a smaller scale since we are more uncertain about the tracklet state compared to
    the detected bounding box position.
    Note that SBIoU reduces to IoU when tracklet confidence is equal to 1.

    \subsection{Using improved similarity measure to find likely objects}
    \label{ssc:similarity}
    DLO confidence boost relies on IoU similarity.
    We propose to replace IoU in equation ~\ref{eq:iouConfBoost} with a more sophisticated similarity measure $S$, which gives:
    \begin{equation}
        \label{eq:likelyConfBoost}
        \hat{c}_{d_i} = \text{max}\big(c_{d_i}, \beta_c\cdot\max_{j}(S(D_i, T_j))\big).
    \end{equation}

    Trivially, we could set $S=\text{SBIoU}$.
    However, we propose to use a richer similarity measure, of which soft BIoU is only a part.

    In ~\cite{stanojevic_boost_2024}, a combination of IoU, shape and Mahalanobis distance similarity was used to construct a "boosted"
    similarity measure which improved association performance.
    If such a similarity measure helps to better distinguish objects, it should also improve the performance of detecting the
    likely objects.

    We define similarity between a detected bounding box $D_i$ and a tracklet $T_j$, $S(D_i, T_j)$, as the average of used similarity measures:
    \begin{equation}
        \label{eq:SimMeasure}
        S(D_i, T_j) = \big(S_1(D_i, T_j) + S_2(D_i, T_j) + \cdots S_p(D_i, T_j)\big)/p.
    \end{equation}
    Intuitively, using an average of multiple similarity measures for the DLO confidence boost means that all similarity measures
    need to "agree" to increase the confidence score of a given detection.
    We use average for simplicity and to give all summands equal weight (note that in equation ~\ref{eq:symBoostEq} weights, i.e. lambdas, do not have to be equal).
    In our implementation, for the DLO confidence boost we use:
    \begin{equation}
        \label{eq:SimMeasureUsed}
        S(D_i, T_j) = \big(\text{SBIoU}(D_i, T_j) + S^{MhD}(D_i, T_j) + S^{shape}(D_i, T_j)\big)/3.
    \end{equation}

    Note that detection-tracklet confidence scores $c_{i,j}$ used to calculate $S^{shape}$
    decrease when detection confidence is low which decreases the $S^{shape}$ and makes it less reliable for the purposes of this work ~\cite{stanojevic_boost_2024}.
    To resolve this issue, we set $c_{i,j}=1$ when using shape similarity for the detection confidence boost.

    \begin{figure}[h!t]
        \centering
        \includegraphics[width=.9\linewidth]{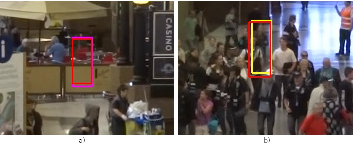}
        \caption{a) Low confidence false positive detection (in red) with high IoU between the tracklet (in purple). Mahalanobis distance similarity between the detection and the tracklet is low,
            and this detection will not be used. b) Low confidence true positive detection (in red) with relatively low IoU between the corresponding tracklet (in yellow).
            However, the average similarity measure when using SBIoU, $S^{shape}$ and $S^{MhD}$ is high enough.}
        \label{fig:s1}
    \end{figure}
    % s1: MOT20-02:156_156_0.85_0.77_0.96_0.82_0.74_0.1599999964237213.
    In figure ~\ref{fig:s1} we show two examples from the MOT20 dataset of low-confidence detections.

    On the left side of the figure, we show false positive detection corresponding to a ghost track.
    The IoU between the detection and the tracklet is high enough (IoU=0.82).
    However, the ghost tracks are static and the corresponding estimated covariance values are low which leads to low
    Mahalanobis distance similarity values even for a small mismatch ($S^{MhD}=0.03$ in the example).
    This results in low average similarity and discarding the false positive detection.

    On the right side of the figure, we show a true positive detection $D_i$ ($c_{d_i}=0.16$) with relatively low IoU between the corresponding tracklet $T_j$ ($\text{IoU}(D_i, T_j)=0.74$).
    However, $\text{SBIoU}(D_i, T_j) = 0.77, S^{MhD}(D_i, T_j) = 0.96$, $S^{shape}(D_i, T_j)=0.82$, and the average similarity (from equation ~\ref{eq:SimMeasure}) is high enough to keep $D_i$ for the association step.

    \subsection{Soft detection confidence boost}
    \label{ssc:softDLO}
    As noted in section ~\ref{sc:limitations}, DLO confidence boost treats equally detections with very low and relatively high confidence scores.
    If the IoU between a given tracklet and a low-confidence detection is high enough (e.g. $0.8$ for the MOT20 dataset),
    the detection's confidence will be increased enough to be used for the association.
    The detection with very low (e.g. $0.05$) and a relatively high confidence score (e.g. $0.35$) require the same high IoU in
    order for their confidence scores to be increased to a value greater than the threshold $\tau$.
    However, the lower confidence detection is more likely to be false-positive and should require higher IoU compared to the
    bounding box with higher confidence score.

    The greater the detection confidence score, the more accurate the corresponding detected bounding box, i.e.
    the greater IoU between the ground-truth and detected bounding box ~\cite{jung2024conftrack}.
    Provided that the tracking module gives accurate predictions, there is also a positive correlation between the detection
    confidence score and the IoU between the tracklet (i.e. predicted bounding box) and the corresponding detected bounding box.
    %There is a positive correlation between the IoU (calculated between the ground-truth and detected bounding boxes)
    % and the detection confidence score ~\cite{jung2024conftrack}.
    % If the tracking module gives accurate predictions, the IoU between a tracklet and the ground-truth should be high and the
    % same correlation applies between the tracklet-detection IoU and the detection confidence.
    IoU between the tracklet and the detection can thus be a measure or an indicator of detection confidence.
    If the IoU between the detected bounding box $D_i$, $D_i \in \{D_1, D_2, \ldots , D_n\}$, and some tracklet is high,
    the detection confidence score $c_{d_i}$ should also be high.
    Following the described logic, for every detected bounding box $D_i$,
    we perform soft detection confidence boost to obtain a new detection confidence score $\hat{c}_{d_i}$ as:
    \begin{equation}
        \label{eq:softConfBoost}
        \hat{c}_{d_i} = \text{max}\Big(c_{d_i}, \alpha \cdot c_{d_i} + (1-\alpha)\cdot \big(\max_{j}(S(D_i, T_j)\big)^q\Big),
    \end{equation}
    where $\alpha \in [0, 1]$ and $q \geq 1$ are hyperparameters and $S$ is any similarity measure.
    We raise $S$ to the power $q$ to have better control over the entire process (all hyperparameter values are discussed in subsection ~\ref{ss:implDetails}).
    Equation ~\ref{eq:softConfBoost} combines original confidence scores with the similarity and solves the problem of
    equal treatment of low and relatively high confidence score detections.

    Following the discussion in subsection ~\ref{ssc:similarity}, $S$ can be defined as in equation ~\ref{eq:SimMeasure}.

\begin{figure}[h!t]
        \centering
        \includegraphics[width=.5\linewidth]{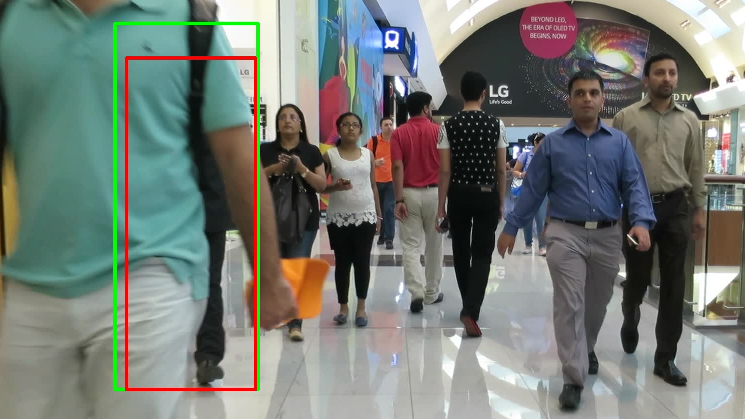}
        \caption{A detected bounding box with confidence (in red) with a confidence score slightly below a threshold, and
        the corresponding tracklet (in green).
        The IoU between the detection and the tracklet is relatively low, but enough for the soft detection confidence
        boost to increase the confidence score above the threshold.}
        \label{fig:sb1}
    \end{figure}

    Figure ~\ref{fig:sb1} shows a frame from MOT17 (sequence 11) and a detected bounding box (in red) with confidence score 0.56 which is slightly below the
    standard threshold 0.6 used for MOT17.
    The IoU between the detection and the corresponding tracklet is relatively low (0.82, which is low compared to the threshold value 0.923 used for MOT17 in ~\cite{stanojevic_boost_2024}).
    However, since the original confidence score is only slightly below the threshold, the imperfect IoU of 0.82 is
    enough to boost the detection confidence above the threshold.

    \subsection{Varying similarity threshold}
    \label{ssc:varyingTh}
    All the similarity measures discussed previously depend upon tracking performance, i.e. the quality of predictions.
    As noted in section ~\ref{sc:limitations}, the quality of predictions reduces with the increase of the number of steps since the last tracklet update.
    Reduced quality of predictions leads to lower IoU value when objects get matched (see figure ~\ref{fig:iou1}) and
    indicates that a single threshold value for DLO confidence boost cannot provide the best results.

    To address the described issue, we propose to use different threshold values when searching for the likely objects.
    The threshold should be higher for the IoU (or any other similarity measure) between a detection and a recently
    updated tracklet.

    Let $\text{last\_update}(T_j)$ be the number of steps since the last update of tracklet $T_j, j \in \{1, 2, \ldots , m\}.$
    We decrease the threshold $\beta_j$ corresponding to a tracklet $T_j$ linearly from the starting value $\beta_{high}$ to the final value $\beta_{low}$,
    and define our varying threshold $\beta_j$ as:
    \begin{equation}
        \label{eq:varyingT}
        \beta_j = \text{max}(\beta_{low}, \beta_{high} - \gamma \cdot (\text{last\_update}(T_j)-1)).
    \end{equation}

    Using the varying similarity threshold, the boosted confidence of the detected bounding box $D_i$, $i \in \{1, 2, \ldots n\}$,
    can be obtained as:
    \begin{equation}
        \hat{c}_{d_i} =
        \begin{cases}
            \text{max}(c_{d_i}, \tau),& \text{if } S(D_i, T_j) \geq \beta_j \:\text{for any } j \in \{1, 2, \ldots, m\},\\
            c_{d_i},              & \text{otherwise,}
        \end{cases}
    \end{equation}
    where $S$ is any similarity measure, and $\tau$ (as noted earlier) threshold for discarding low-confidence detections.

\begin{figure}[h!t]
        \centering
        \includegraphics[width=.5\linewidth]{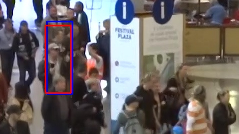}
        \caption{A detected bounding box $D_i$ (in red) with an original confidence score of 0.35. A predicted bounding box (in blue) after not matching the tracklet $T_j$ for 27 frames.
        For the given tracklet, threhold $\beta_j=0.8$. Since $\text{IoU}(D_i,T_j)=0.83\geq \beta_j$, the $c_{d_i}$ is increased, and $D_i$ will not be discarded.}
        \label{fig:vt1}
    \end{figure}

    An example of low-confidence detected bounding box $D_i$ with high enough IoU between some tracklet $T_j$ is shown in figure ~\ref{fig:vt1}.

    \subsection{Combining the proposed methods}
    \label{ssc:methodAll}
    All proposed additions are mutually independent and can be used separately or jointly.
    Algorithm ~\ref{alg:cap} shows the combined usage of all proposed elements and can be used as
    a replacement for the DLO confidence boost used in BoostTrack ~\cite{stanojevic_boost_2024}.
    Parameters $useS$, $useSB$ and $useVT$ control whether we should use improved similarity measure,
    soft detection confidence boost and varying threshold, respectively.

    \begin{algorithm}[ht]
        % \centering
        \caption{Improved detection confidence boost}\label{alg:cap}
        \begin{algorithmic}[1]
            \Procedure{IDCBoost}{$D, T, useS, useSB, useVT$}
                \Comment{$D = \{D_1, D_2, \ldots D_n\}$, $T = \{T_1, T_2 \ldots , T_m\}$}
                \If{useS}
                    \State S := compute\_similarity($D$, $T$) \Comment{Using equation \ref{eq:SimMeasureUsed}.}
                \Else
                    \State S := IoU($D$, $T$)
                \EndIf

                \If{not useSB and not useVT}
                    \For{ each i}
                        \State $\hat{c}_{d_i} := \text{max}\big(c_{d_i}, \beta_c\cdot\max_{j}(S(D_i, T_j))\big)$.
                        \Comment{Applying equation ~\ref{eq:likelyConfBoost}}
                    \EndFor
                \Else
                    \If{use SB}
                        \For{ each i}
                            \State $c_{d_i} := \text{max}\Big(c_{d_i}, \alpha \cdot c_{d_i} + (1-\alpha)\cdot \big(\max_{j}(S(D_i, T_j)\big)^q\Big)$
                            \Comment{Applying equation ~\ref{eq:softConfBoost}}
                        \EndFor
                    \EndIf
                    \If{use VT}
                        \For{ each (i, j)}
                            \State $beta_j$ := compute\_threshold($T_j, \gamma, \beta_{low}, \beta_{high}$) \Comment{Using equation \ref{eq:varyingT}.}
                            \If{$S(D_i, T_j) \geq beta_j$}
                                \State $c_{d_i}$ := $\text{max}(c_{d_i}, \tau)$
                                \State \textbf{break}
                            \EndIf
                        \EndFor
                    \EndIf
                \EndIf

                \State \textbf{return} $c_{d_1}, c_{d_2}, \ldots , c_{d_n}$
                \Comment{Outputs boosted confidence scores $\hat{c}_{d_1}, \hat{c}_{d_2}, \ldots, \hat{c}_{d_n}.$}
            \EndProcedure
        \end{algorithmic}
    \end{algorithm}

    \section{Experiments and results}
    \label{sec:experiments}
    \subsection{Datasets and metrics}
    \label{ssc:metrics}
    \textbf{Datasets.} We perform experiments and evaluation of our method on standard MOT benchmark datasets: MOT17 ~\cite{milan_mot16_2016}
    and MOT20 ~\cite{MOTChallenge20}.
    MOT17 contains pedestrian videos filmed with static and moving camera.
    The training set consists of $7$ sequences and $5316$ frames in total (frame rate ranges from $14$ to $30$ depending on the video),
    while the test set contains a continuation of the same sequences and has $5919$ frames in total.

    MOT20 consists of 8 sequences of crowded scenes with changing lighting conditions filmed at 25 FPS.
    The training set contains 4 sequences ($8931$ frames) and the test set consists of remaining sequences ($4479$ frames).

    As in previous works (e.g. ~\cite{aharon_bot-sort_2022, zhang_bytetrack_2022, maggiolino_deep_2023}), we
    use a custom detector instead of the provided dataset detections and
    perform experiments under private detection protocol.
    We use the second half of each training sequence as a validation set.

    \textbf{Metrics.} We use standard metrics to assess the performance of our method.
    Namely, we use:
    \begin{itemize}
        \item Multi-Object Tracking Accuracy (MOTA) metric ~\cite{bernardin2008evaluating}, which penalizes false positive and false negative detections
        and is primarily used to evaluate detection performance.
        \item IDF1 ~\cite{ristani2016performance}, which is primarily used as a measure of
        association performance.
        \item Higher Order Tracking Accuracy (HOTA) ~\cite{luiten_hota_2021}, which combines localization accuracy,
        association and tracking performance and tends to assess the entire MOT performance.
    \end{itemize}
    In addition to the mentioned metrics, in the ablation study, we also monitor the number of IDs used and the number of identity switches (IDSW).
    We monitor IDs because detection confidence boosting can result in an increased number of IDs.
    IDWS is important because new detections should not cause additional IDSWs.

    \subsection{Implementation details}
    \label{ss:implDetails}
    \textbf{MOT and BoostTrack specific settings.} We extend the work done in ~\cite{stanojevic_boost_2024} and use the same additional components and settings.
    We provide a brief overview of the most important components used\footnote{We instruct readers to ~\cite{stanojevic_boost_2024}
    for more information on minor implementation details.}.
    Namely, we use YOLOX-X ~\cite{ge2021yolox} as the detector with weights from ~\cite{zhang_bytetrack_2022};
    we apply Enhanced correlation coefficient maximization from ~\cite{evangelidis_parametric_2008} for camera motion compensation (and use the implementation from ~\cite{du_strongsort_2023});
    we use FastReID ~\cite{he_fastreid_2020} for computing visual embedding (used for calculating visual appearance similarity);
    for postprocessing of the results, we use gradient boosting interpolation (GBI) from ~\cite{zeng2023nct}.
    We calculate shape mismatch (used in $S^{shape}$) between a tracklet $T_j$ and a detection $D_i$, $ds_{i,j}$ as:
    \begin{equation}
        ds_{i,j} = \frac{|D_i^w - T_j^w|+|D_i^h - T_j^h|}{\text{max}(D_i^w, T_j^w)}.
    \end{equation}

    \textbf{BoostTrack++ specific settings.}
    We run a grid search to find optimal parameters $q$ and $\alpha$ (used for soft detection confidence boost introduced in subsection ~\ref{ssc:softDLO}).
    Specifically, we tested settings $(q, \alpha) \in \{1, 1.25, 1.5, 1.75, 2\} \times \{0, 0.05, 0.1, \ldots , 0.95, 1\}$
    on MOT17 validation set (we used $S = \text{IoU}$ for simplicity) and choose $q = 1.5, \alpha = 0.65$ as the best trade-off
    between different metrics and different baseline settings (e.g. whether we use camera motion compensation, postprocessing, appearance similarity).

    Based on empirical results displayed in figure ~\ref{fig:iou1}, as our varying similarity threshold setting we used
    $\beta_{high} = 0.95,$ which we reduce to $\beta_{low} = 0.8$ over 20 frames ($\gamma = (\beta_{high}-\beta_{low})/20 = 0.0075$).

    \textbf{Software.} We build our code on top of code from ~\cite{stanojevic_boost_2024},
    which uses codes from ~\cite{maggiolino_deep_2023, bewley_simple_2016, zhang_bytetrack_2022, du_strongsort_2023, zeng2023nct}.

    \textbf{Hardware.} All experiments are performed on the desktop with
    13th Gen Intel(R) Core(TM) i9-13900K CPU and NVIDIA GeForce RTX 3080 GPU.

    We use TrackEval ~\cite{luiten2020trackeval} to evaluate the results on validation sets.
    The results on the test sets are evaluated on the official MOT Challenge server~\cite{MOTChallengeURL}.

    \subsection{Ablation study}
    \label{ssc:Ablation}
    The goal of the proposed method is to replace or improve the performance of the DLO confidence boost technique
    used in ~\cite{stanojevic_boost_2024}.
    Ideally, our method should not increase the number of used IDs compared to any baseline which does not apply DLO
    (because it should only boost the confidence score of the detections which correspond to existing objects).
    Furthermore, a successful DLO confidence boost should improve the overall tracking performance:
    not only the MOTA score which penalizes usage of false positive detections,
    but the values of HOTA, IDF1 and IDWS metrics should also improve because the quality of selected bounding boxes
    is crucial for the successful matching.

    \subsubsection{Influence of different similarity measures.}
    Since the DLO confidence boost resulted in substantial performance improvement on the MOT20, but also in an increased number of
    IDs and IDWSs, we study the effect of various similarity measures from equation ~\ref{eq:SimMeasureUsed} on MOT20 validation set.
    We performed experiments on three baselines: SORT+DLO, BoostTrack and BoostTrack+.
    Table ~\ref{tb:ablationS} shows the results\footnote{We consider the best IDs and IDWS values that are the closest to non-DLO baseline.}
    of using SORT+DLO and BoostTrack+ as baselines (as the most basic and the more advanced methods, respectively).
    We display the results of using the BoostTrack baseline in ~\ref{sec:appA}.

    The first two rows of table ~\ref{tb:ablationS} are the baselines.
    The first row shows results of the baseline method without DLO confidence boost, while the second uses a simple IoU based
    DLO confidence boost which we try to improve (see equation ~\ref{eq:iouConfBoost}).
    A successful DLO method should keep IDs and IDSW values as close to the values from the first row,
    and at the same time improve the HOTA, MOTA and IDF1 values from the second row.

    \addtolength{\tabcolsep}{-3.5pt}
    \begin{table}[h!t]
        \center
        \caption{The effect of various similarity measures used for DLO confidence boost on the MOT20 validation set (best in bold).}
        \begin{tabular}{cccc|ccccc|ccccc}
            \toprule
            \multicolumn{4}{c|}{Setting} & \multicolumn{5}{c|}{SORT baseline} & \multicolumn{5}{c}{BoostTrack+ baseline} \\%\cmidrule{1-11}
            DLO & SBIoU & $S^{MhD}$ & $S^{shape}$ & HOTA & MOTA & IDF1 & IDSW & IDs & HOTA & MOTA & IDF1 & IDSW & IDs \\
            \midrule
            \xmark & \xmark & \xmark & \xmark & 56.65 & 69.91 & 73.6 & 1127 & 1894 & 62.59 & 77.16 & 79.29 & 730 & 1852 \\
            \cmark & \xmark & \xmark & \xmark & \textbf{58.58} & 73.28 & 75.09 & 1259 & 2045 & 62.58 & 77.7 & 78.93 & 794 & 2019 \\
            \cmark & \xmark & \xmark & \cmark & 57.55 & \textbf{73.4} & 72.59 & 2008 & 2729 & 61.99 & 75.63 & 77.51 & 1081 & 2736 \\
            \cmark & \xmark & \cmark & \xmark & 57.76 & 71.3 & 74.87 & 1188 & 1932 & 62.79 & 77.71 & 79.37 &  739 & 1917 \\
            \cmark & \xmark & \cmark & \cmark & 57.73 & 71.33 & 74.81 & 1186 & \textbf{1926} & 62.8 & 77.75 & 79.35 & \textbf{723} & \textbf{1914} \\
            \cmark & \cmark & \xmark & \xmark & 58.52 & 73.19 & 74.94 & 1228 & 1981 & 62.53 & \textbf{77.84} & 78.77 & 780 & 1964 \\
            \cmark & \cmark & \xmark & \cmark & 58.5 & 73.22 & 74.97 & 1273 & 2040 & 62.63 & 77.7 & 78.82 & 782 & 2038 \\
            \cmark & \cmark & \cmark & \xmark & 57.87 & 71.3 & 75.08 & \textbf{1158} & 1934 & \textbf{62.84} & 77.78 & \textbf{79.49} & 736 & 1928 \\
            \cmark & \cmark & \cmark & \cmark & 57.95 & 71.38 & \textbf{75.2} & 1162 & 1929 & 62.82 & 77.8 & 79.42 & 729 & 1918 \\
            \bottomrule
        \end{tabular}
        \label{tb:ablationS}
    \end{table}
    \addtolength{\tabcolsep}{3.5pt}

    As results from the table show, using an average of $\text{SBIoU}$, $S^{shape}$ and $S^{MhD}$ offers the best trade-off between
    the various metrics.
    It substantially reduces IDs and IDSW values compared to the DLO baseline, while increasing HOTA, MOTA and IDF1 values
    (the only exception is MOTA score when using the SORT baseline).

    \subsubsection{Influence of other components}
    We study the influence of proposed components on BoostTrack+ baseline on MOT17 and MOT20 validation sets.
    In ~\ref{sec:appB} we show the results of the ablation study where we used SORT ~\cite{bewley_simple_2016} and BoostTrack as baselines.

    Table ~\ref{tb:ablationBTPlus} shows the influence of adding each of the proposed components to the BoostTrack+ baseline:
    namely, new similarity measure for the confidence boost (S column), soft (detection confidence) boost (SB column), and
    using varying threshold (VT column) for the DLO confidence boost.
    Note, when we use SB or VT, and not S, the similarity used in SB and VT is IoU.
    For more details, see algorithm ~\ref{alg:cap}.
    Again, in the first row, we show results of the baseline method without any DLO confidence boost, while the second row
    shows results of the original DLO confidence boost from BoostTrack.

    \addtolength{\tabcolsep}{-3.5pt}
    \begin{table}[h!t]
        \center
        \caption{Ablation study on the MOT17 and MOT20 validation sets for different additional components (best in bold). Components are added to the BoostTrack+ baseline.}
        \begin{tabular}{cccc|ccccc|ccccc}
            \toprule
            \multicolumn{4}{c|}{Setting} & \multicolumn{5}{c|}{MOT17} & \multicolumn{5}{c}{MOT20} \\
            DLO & S & SB & VT & HOTA & MOTA & IDF1 & IDSW & IDs & HOTA & MOTA  & IDF1 & IDSW & IDs \\
            \midrule
            \xmark & \xmark & \xmark & \xmark & 72.17 & 81.02 & 84.86 & 79 & 388
            & 62.59 & 77.16 & 79.29 & 730 & 1852 \\
            \cmark & \xmark & \xmark & \xmark & \textbf{72.41} & 80.95 & 85.09 & 83 & 402 &
            62.58 & 77.7 & 78.93 & 794 & 2019 \\
            \cmark & \xmark & \xmark & \cmark & 72.13 & 80.88 & 84.76 & 84 & {\bf 395} &
            62.8 & 77.3 & 79.56 & {\bf 712} & 1852 \\
            \cmark & \xmark & \cmark & \xmark & 72.05 & 80.67 & 84.91 & {\bf 80} &  402 &
            62.86 & 77.84 & 79.23 & 812 & 2043 \\
            \cmark & \xmark & \cmark & \cmark & 72.02 & 80.58 & 85.08 & 81 & 412 &
            62.82 & 77.86 & 79.16 & 813 & 2042 \\
            \cmark & \cmark & \xmark & \xmark &  71.88 & 80.96 & 84.82 & 89 & 398 &
            62.82 & 77.8 & 79.42 & 729 & 1918  \\
            \cmark & \cmark & \xmark & \cmark & 72.17 & 81.15 & 84.75 & 83 & 398 &
            62.75 & 77.35 & 79.44 & 722 & {\bf 1848} \\
            \cmark & \cmark & \cmark & \xmark & 72.24 & 81.23 & 85.44 & 81 & 396 &
            \textbf{63.2} & \textbf{78.03} & \textbf{80.05} & 734 & 1912 \\
            \cmark & \cmark & \cmark & \cmark & 72.22 & \textbf{81.33} & \textbf{85.45} & 84 & 396 &
            63.16 & 78.02 & 79.97 & 732 & 1906 \\

            \bottomrule
        \end{tabular}
        \label{tb:ablationBTPlus}
    \end{table}
    \addtolength{\tabcolsep}{3.5pt}

    The results from table ~\ref{tb:ablationBTPlus} show that using SB and VT with IoU as a similarity measure (without the proposed average similarity)
    increases the number of IDs even more than baseline DLO.
    However, combined with S, we get the best of both worlds:
    we get a substantial improvement in tracking performance, outperforming the original DLO confidence boost, while at the same time
    only slightly increasing the number of IDs and IDSWs:
    +5 IDSWs (6.3\%), +8 IDs (2\%) on the MOT17 and +2 IDSWs (0.3\% ), +54 IDs (2.9\%) on the MOT20.

    \subsection{Comparison with other methods}
    To evaluate our method on MOT17 ~\cite{milan_mot16_2016} and MOT20 ~\cite{MOTChallenge20} test sets (under private detection protocol), we used all proposed additions combined
    (S+SB+VT setting).
    The results on the test sets are provided in the table ~\ref{tb:mottest}.
    We mark offline methods with '*'\footnote{
        Note that our method, as most "online" methods (e.g. ~\cite{zhang_bytetrack_2022, aharon_bot-sort_2022, stadler_improved_2023, maggiolino_deep_2023, jung2024conftrack}),
        applies postprocessing to the results (which are obtained online), and could be best described as "semi-online".
        However, the official MOT Challenge ~\cite{MOTChallengeURL} distinguishes only offline and online methods, and methods like ours are considered online.
    }.

    \begin{table}[h!t]
        \caption{Comparison with other MOT methods on the MOT17 test set (best in bold). Offline methods are marked with '*'.}
        \begin{tabular}{c|cccc|cccc}
            \toprule
            \multirow{2}{*}{Method} & \multicolumn{4}{c}{MOT17} & \multicolumn{4}{c}{MOT20}\\
            & HOTA & MOTA & IDF1 & IDSW & HOTA & MOTA & IDF1 & IDSW \\
            \midrule
            FairMOT ~\cite{zhang_fairmot_2021} & 59.3 & 73.7 & 72.3  & 3303 & 54.6 & 61.8 & 67.3 & 5243\\
            MOTR ~\cite{zeng2022motr} & 62.0 & 78.6 & 75.0 & 2619 & / & / & / & / \\
            ByteTrack ~\cite{zhang_bytetrack_2022} & 63.1 & 80.3 & 77.3 & 2196  & 61.3 & 77.8 & 75.2 & 1223\\
            QuoVadis ~\cite{dendorfer2022quo} & 63.1 & 80.3 & 77.7 & 2103 & 61.5 & 77.8 & 75.7  & 1187\\
            BPMTrack ~\cite{gao2024bpmtrack} & 63.6 & 81.3 & 78.1 & 2010 & 62.3 & 78.3 & 76.7 & 1314 \\
            SuppTrack* ~\cite{zhang2023handling} & / & / & / & / & 61.9 & 78.2 & 75.5 & 1325 \\
            UTM ~\cite{you2023utm} & 64.0 & 81.8 & 78.7 & 1431 & 62.5 & 78.2 & 76.9 & 1228 \\
            FineTrack ~\cite{ren2023focus} & 64.3 & 80.0 & 79.5 & 1272 & 63.6 & 77.9 & 79.0 & 980\\
            StrongSORT++ ~\cite{du_strongsort_2023} & 64.4 & 79.6 & 79.5 & 1194 & 62.6 & 73.8 & 77.0 & 770 \\
            BASE* ~\cite{larsen2023base} & 64.5 & 81.9 & 78.6 & 1281 & 63.5 & 78.2 & 77.6 & 984\\
            Deep OC-SORT ~\cite{maggiolino_deep_2023} & 64.9 & 79.4 & 80.6 & 1023 & 63.9 & 75.6 & 79.2 & 779\\
            BoT\_SORT ~\cite{aharon_bot-sort_2022} & 65.0 & 80.5 & 80.2 & 1212 & 63.3 & 77.8 & 77.5 & 1313\\
            SparseTrack ~\cite{liu2023sparsetrack} & 65.1 & 81.0 & 80.1 & 1170 & 63.5 & 78.1 & 77.6 & 1120\\
            MotionTrack ~\cite{qin2023motiontrack} & 65.1 & 81.1 & 80.1 & 1140 & 62.8 & 78.0 & 76.5  & 1165\\
            LG-Track ~\cite{meng2023localization} & 65.4 & 81.4 & 80.4 & 1125 & 63.4 & 77.8 &   77.4 & 1161\\
            StrongTBD ~\cite{stadler2023detailed} & 65.6 & 81.6 & 80.8 & 954 & 63.6 & 78.0 & 77.0 & 1101\\
            C-BIoU ~\cite{yang_hard_2023} & 66.0 & \textbf{82.8} & 82.5 & 1194 & / & / & / & / \\
            PIA2 ~\cite{stadler2023past} & 66.0 & 82.2 & 81.1 & 1026 & 64.7 & 78.5 & 79.0 & 1023\\
            ImprAsso ~\cite{stadler_improved_2023} & 66.4  & 82.2 & 82.1 & \textbf{924} & 64.6 & \textbf{78.6} & 78.8 & 992\\
            SUSHI* ~\cite{cetintas2023unifying} & 66.5 & 81.1 & 83.1  & 1149 & 64.3 & 74.3 & 79.8 & 706\\
            ConfTrack ~\cite{jung2024conftrack} & 65.4 & 80.0 & 81.2 & 1155 & 64.8 & 77.2 & 80.2 & \textbf{702} \\
            BoostTrack+ ~\cite{stanojevic_boost_2024} & 66.4 & 80.6 & 81.8 & 1086 & 66.2 & 77.2 & 81.5 & 827\\
            CoNo-Link* ~\cite{Gao_Xu_Li_Wang_Gao_2024} & \textbf{67.1} & 82.7 & \textbf{83.7} & 1092 & 65.9 & 77.5 & 81.8 & 956 \\
            \midrule
            BoostTrack++ (\textbf{ours}) & 66.6 & 80.7 & 82.2 & 1062 & \textbf{66.4} & 77.7 & \textbf{82.0} & 762 \\
            \bottomrule
        \end{tabular}
        \label{tb:mottest}
    \end{table}

    Our BoostTrack++ method achieves improvement compared to the BoostTrack+ on the MOT17 test set: +0.2 HOTA, +0.1 MOTA, +0.4 IDF1, -24 IDSWs.
    On more challenging MOT20, BoostTrack++ shows slightly better improvement compared to the BoostTrack+: +0.2 HOTA, +0.5 MOTA, +0.5 IDF1, -65 IDSWs.
    Improvement in MOTA indicates the increase in the use of true positive detections, which improves HOTA and IDF1 scores.

    Compared to the online trackers, our method ranks first in HOTA score (66.6)
    on the MOT17 test set, while on the MOT20, our method ranks first in HOTA (66.4) and IDF1 (82.0) among both online and offline methods.

    \section{Conclusions}
    \label{sec:conclusion}
    In this paper, we identified the drawbacks of the DLO confidence boost introduced in BoostTrack and proposed a method to mitigate the identified issues.
    Our goal was to utilize the benefits of DLO confidence boost but avoid its drawbacks - namely, causing IDSWs and introducing new IDs.
    To this end, we proposed a novel soft BIoU similarity measure and three plug-and-play additions, each of which attempts to
    provide better control and utilize richer tracklet and detection information to improve the selection of true
    positive detected bounding boxes.

    Using our methods with BoostTrack+ baseline, our BoostTrack++ method ranks first in HOTA and IDF1 metrics
    on the MOT20 dataset and achieves comparable to the state of the art results on the MOT17 dataset.

    However, the achieved MOTA score is still relatively low, indicating the need for even better algorithms for
    selecting true positive detections in a one-stage TBD MOT paradigm.

    \appendix

    \section{List of abbreviations}
    \label{sec:appAbr}
    In table ~\ref{tb:abbr}, we show the list of all abbreviations used in the paper.

    \addtolength{\tabcolsep}{-2pt}
    \begin{table}[!htbp]
        \small
        \center
        \caption{List of abbreviations} % and acronyms}
        \begin{tabular}{cc}
            \toprule
            Abbreviation & Definition \\
            \midrule
            TBD MOT & Tracking by detection multiple object tracking\\
            IoU & Intersection over union \\
            BIoU & Buffered IoU from ~\cite{yang_hard_2023}\\
            SBIoU & Soft BIoU introduced in subsection ~\ref{ssc:sbiou} \\
            DLO & Detection of likely objects used to boost confidence scores in ~\cite{stanojevic_boost_2024}\\
            IDSW & Identity switch\\
            S (ablation study) & Similarity measure from equation ~\ref{eq:SimMeasureUsed}\\
            SB (ablation study) & Soft (detection confidence) boost from subsection ~\ref{ssc:softDLO}\\
            VT (ablation study) & Varying similarity threshold from subsection ~\ref{ssc:varyingTh}  \\
            \bottomrule
        \end{tabular}
        \label{tb:abbr}
    \end{table}
    \addtolength{\tabcolsep}{2pt}

    \section{Influence of different similarity measures}
    \label{sec:appA}
    In table ~\ref{tb:ablationSBT} we show the results of applying various similarity measures instead of IoU for the DLO confidence boost.
    We used BoostTrack as the baseline method.
    As with SORT or BoostTrack+ baselines (displayed in table ~\ref{tb:ablationS}), using the average of all three similarity measures provides the best
    trade-off between introducing additional identities and overall tracking performance.
    \begin{table}[!htbp]
        \center
        \caption{The effect of various similarity measures used for the DLO confidence boost on the MOT20 validation set (best in bold).}
        \begin{tabular}{cccc|ccccc}
            \toprule
            \multicolumn{4}{c|}{Setting} & \multicolumn{5}{c}{BoostTrack baseline} \\%\cmidrule{1-11}
            DLO & SBIoU & $S^{MhD}$ & $S^{shape}$ & HOTA & MOTA & IDF1 & IDSW & IDs \\
            \midrule
            \xmark & \xmark & \xmark & \xmark & 61.39 & 77.02 & 77.23 & 803 & 1867 \\
            \cmark & \xmark & \xmark & \xmark & 61.74 & 77.46 & 77.45 & 898 & 2008 \\
            \cmark & \xmark & \xmark & \cmark & 59.64 & 75.34 & 73.85 & 1252 & 2637 \\
            \cmark & \xmark & \cmark & \xmark & 61.7 & 77.38 & 77.52 & 856 & 1911 \\
            \cmark & \xmark & \cmark & \cmark & 61.69 & 77.52 & 77.57 & 862 & 1911 \\
            \cmark & \cmark & \xmark & \xmark & 61.74 & \textbf{77.65} & 77.58 & 865 & 1949 \\
            \cmark & \cmark & \xmark & \cmark & 61.17 & 77.5 & 76.63 & 915 & 2026 \\
            \cmark & \cmark & \cmark & \xmark & 61.8 & 77.6 & 77.7 & 834 & 1917 \\
            \cmark & \cmark & \cmark & \cmark & \textbf{61.88} &  77.57 & \textbf{77.89} & \textbf{825} & \textbf{1902} \\
            \bottomrule
        \end{tabular}
        \label{tb:ablationSBT}
    \end{table}

    \section{Ablation study on SORT and BoostTrack}
    \label{sec:appB}
    % We also performed ablation study using SORT and BoostTrack as baseline methods.
    % The results are shown in tables ~\ref{tb:ablationSORT} (SORT baseline) and ~\ref{tb:ablationBT} (BoostTrack baseline).
    To show the robustness of the proposed components, we performed additional ablation experiments using SORT and BoostTrack as baseline methods.
    Table ~\ref{tb:ablationSORT} shows the results of various settings on MOT20 validation set.
    Note that, ideally, our method should not introduce new IDs and IDSWs, and it should at the same time improve overall tracking performance.
    Using only VT with BoostTrack as the baseline provides a good example of trade-offs required.
    Adding VT even reduced IDs and IDWSs by 2.
    However, the overall tracking performance is only slightly improved by adding VT only,
    while S or S+SB+VT settings provide a better trade-off between various metrics and the number of additional IDs and IDSWs.

    \addtolength{\tabcolsep}{-2pt}
    \begin{table}[!htbp]
        \center
        \caption{Ablation study on the MOT20 validation sets for different additional components (best in bold).}
        \begin{tabular}{cccc|ccccc|ccccc}
            \toprule
            \multicolumn{4}{c|}{Setting} & \multicolumn{5}{c|}{SORT baseline} & \multicolumn{5}{c}{BoostTrack baseline} \\%\cmidrule{1-11}
            DLO & S & SB & VT & HOTA & MOTA & IDF1 & IDSW & IDs & HOTA & MOTA  & IDF1 & IDSW & IDs \\
            \midrule
            \xmark & \xmark & \xmark  & \xmark & 56.65 & 69.91 & 73.6 & 1127 & 1894 &
            61.39 & 77.02 & 77.23 & 803 & 1867 \\
            \cmark & \xmark & \xmark & \xmark & 58.58 & {\bf 73.28} & 75.09 & 1259 & 2045 &
            61.74 & 77.46 & 77.45 & 898 & 2008 \\
            \cmark & \xmark & \xmark & \cmark & 56.62 & 70.13 & 73.6 & {\bf 1130} & 1901 &
            61.53 & 77.2 & 77.46 & {\bf 801} & {\bf 1865}\\
            \cmark & \xmark & \cmark & \xmark & \textbf{58.07} & 73.06 & 74.55 & 1315 & 2059 &
            61.29 & 77.61 & 76.88 & 902 & 2052 \\
            \cmark & \xmark & \cmark & \cmark & 58.03 & 73.06 & 75.0 & 1325 & 2060 &
            61.25 & 77.59 & 76.84 & 904 & 2048 \\
            \cmark & \cmark & \xmark & \xmark & 57.95 & 71.38 & {\bf75.2} & 1162 & 1929 &
                {\bf 61.88} & 77.57 & {\bf 77.89} & 825 & 1902\\
            \cmark & \cmark & \xmark & \cmark & 56.9 & 70.41 & 73.97 & 1148 & {\bf 1891} &
            61.54 & 77.23 & 77.48 & 819 & 1862 \\
            \cmark & \cmark & \cmark & \xmark & 57.61 & 71.65 & 74.53 & 1192 & 1933 &
            61.57 & 77.67 &  77.43 & 866 & 1901 \\
            \cmark & \cmark & \cmark & \cmark & 57.65 & 71.64 & 74.6 & 1200 & 1935 &
            61.56 & \textbf{77.71} & 77.38 & 864 & 1895 \\
            \bottomrule
        \end{tabular}
        \label{tb:ablationSORT}
    \end{table}
    \addtolength{\tabcolsep}{2pt}
    % \clearpage
    
% \bibliographystyle{} 

\bibliographystyle{unsrt}  
\bibliography{references}

% %\bibliography{references}  %%% Remove comment to use the external .bib file (using bibtex).
% %%% and comment out the ``thebibliography'' section.

% %%% Comment out this section when you \bibliography{references} is enabled.

% George Kour and Raid Saabne.
% \newblock Real-time segmentation of on-line handwritten arabic script.
% \newblock In {\em Frontiers in Handwriting Recognition (ICFHR), 2014 14th
%   International Conference on}, pages 417--422. IEEE, 2014.

% George Kour and Raid Saabne.
% \newblock Fast classification of handwritten on-line arabic characters.
% \newblock In {\em Soft Computing and Pattern Recognition (SoCPaR), 2014 6th
%   International Conference of}, pages 312--318. IEEE, 2014.

% \bibitem{hadash2018estimate}
% Guy Hadash, Einat Kermany, Boaz Carmeli, Ofer Lavi, George Kour, and Alon
% \newblock Estimate and replace: A novel approach to integrating deep neural
%   networks with existing applications.
% \newblock {\em arXiv preprint arXiv:1804.09028}, 2018.

% \end{thebibliography}

\end{document}